\pdfoutput=1

\documentclass[11pt]{article}

\usepackage[final]{acl}

\usepackage{times}
\usepackage{latexsym}

\usepackage[T1]{fontenc}

\usepackage[utf8]{inputenc}

\usepackage{microtype}

\usepackage{inconsolata}

\usepackage{graphicx}

\usepackage[colorinlistoftodos,prependcaption,textsize=tiny]{todonotes}
\usepackage{booktabs}
\usepackage{amsmath}
\usepackage{tcolorbox}
\usepackage{paralist}

\title{CLEAR: A Comprehensive Linguistic Evaluation of Argument Rewriting by Large Language Models}

\author{Thomas Huber \\
  University of St. Gallen, Switzerland \\
  \texttt{thomas.huber@unisg.ch} \\\And
  Christina Niklaus \\
  University of St. Gallen, Switzerland \\
  \texttt{christina.niklaus@unisg.ch} \\}

\begin{document}
\maketitle
\begin{abstract}
While LLMs have been extensively studied on general text generation tasks, there is less research on text rewriting, a task related to general text generation, and particularly on the behavior of models on this task. In this paper we analyze what changes LLMs make in a text rewriting setting. We focus specifically on argumentative texts and their improvement, a task named Argument Improvement (ArgImp). We present CLEAR: an evaluation pipeline consisting of 57 metrics 
mapped to four linguistic levels: lexical, syntactic, semantic and pragmatic. This pipeline is used to examine the qualities of LLM-rewritten arguments on a broad set of argumentation corpora and compare the behavior of different LLMs on this task and analyze the behavior of different LLMs on this task in terms of linguistic levels. By taking all four linguistic levels into consideration, we find that the models perform ArgImp by shortening the texts while simultaneously increasing average word length and merging sentences. Overall we note an increase in the persuasion and coherence dimensions. 
\end{abstract}


\section{Introduction}
Text rewriting is an important task in Natural Language Processing, with applications in style transfer \cite{Fu_Tan_Peng_Zhao_Yan_2018,10.1145/3544903.3544906,reif-etal-2022-recipe,riley-etal-2021-textsettr}, paraphrase generation \cite{zhou-bhat-2021-paraphrase,li-etal-2018-paraphrase}, and text simplification \cite{shardlow2014survey,saggion2017automatic,alva-manchego-etal-2020-asset}, among others. It can be seen as a form of controllable text generation \cite{10.1145/3617680}, where a given text is modified based on specific requirements, such as improving readability, accuracy, or suitability for a particular context \cite{dou-etal-2024-automatic}. Recent advancements in large language models (LLMs) have shown promising performance on a wide range of text generation tasks, allowing them to refine text based on natural language instructions to produce high-quality rewrites \cite{53290}.

A relevant but underexplored application of text rewriting is the task of ArgImp, i.e. 
rephrasing an argumentative text, with the objective of enhancing its overall quality.
Arguments can be refined through various linguistic modifications, including lexical, syntactic, semantic, and pragmatic changes. LLMs have been increasingly studied in the domain of Computational Argumentation, with recent works showcasing their capabilities in the tasks of Argument Mining \cite{chen-etal-2024-exploring-potential,abkenar2024assessingopensourcelargelanguage}, Argument Generation \cite{chen-etal-2024-exploring-potential,eskandari-miandoab-sarathy-2024-lets,kao-yen-2024-magic}, and Argument Quality Assessment \cite{wachsmuth-etal-2024-argument,10.1007/978-3-031-63536-6_8}.

\begin{figure*}
    \centering
    \includegraphics[width=1\linewidth]{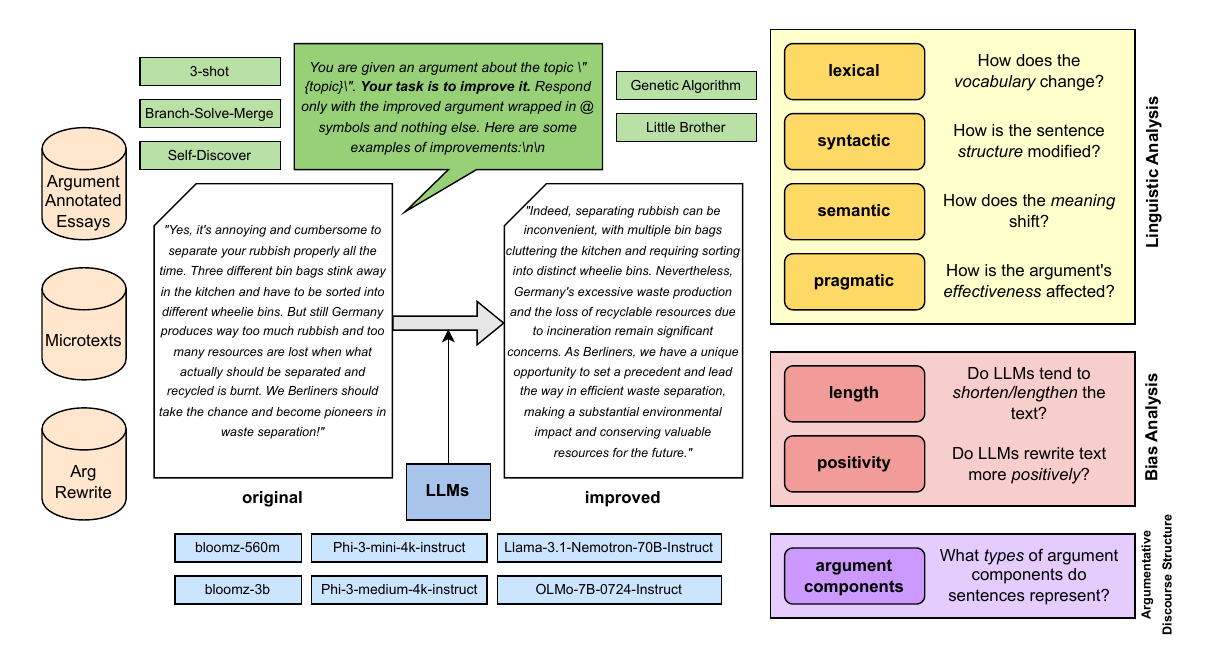}
    \caption{Overview of our experimental setup for the task of ArgImp. We evaluate the quality of argumentative texts rewritten by LLMs prompted for improvement. We apply six models across five datasets (each revision of the ArgRewrite corpus is treated as a distinct dataset). The evaluation spans four linguistic levels, examines two types of biases, and compares the argumentative discourse structure of the original and improved texts.}
    \label{fig:overview}
\end{figure*}

This work aims to bridge this gap by investigating the linguistic transformations performed by LLMs when prompted to improve an argumentative text. Specifically, we analyze how these models alter texts at four distinct linguistic levels: word choice \textit{(lexical)}, sentence structure \textit{(syntactic)}, meaning shifts \textit{(semantic)}, and rhetorical effectiveness \textit{(pragmatic)}. By systematically categorizing and evaluating these modifications, we aim to better understand the role of LLMs in ArgImp and their potential for enhancing argumentative writing (see Figure \ref{fig:overview}).

LLMs are known to exhibit biases in text generation settings \cite{oketunji2023large}. Due to a lack of research investigating LLMs in an ArgImp scenario, it is not clear what, if any, biases they exhibit in this setting. 
A so called verbosity bias can be observed in LLMs \cite{chen-etal-2024-humans,10.5555/3666122.3668142}, as well as a positivity bias \cite{Palmer02092023,buhnila-etal-2025-chain,markowitz2024linguistic}. These biases are of particular interest in an ArgImp setting. The models may consider longer texts as better, and thus make less changes, or shift the tone and inadvertently change the meaning of the original argument. For this reason we include an investigation into length and positivity biases.

To investigate the behavior of LLMs in an ArgImp setting we have created an evaluation pipeline consisting of 57 metrics commonly used in natural language generation (NLG), named CLEAR\footnote{Comprehensive Linguistic Evaluation for Argument Rewriting}. These include scores that measure lexical, syntactic, semantic and pragmatic aspects of the texts. The focus of our work is on analyzing what changes the models make when used in an ArgImp setting, but the pipeline and approach are applicable to other text generation tasks as well. We applied five different prompting techniques to make LLMs write improved versions of arguments from the Microtexts \cite{peldszus2015annotated} (both English and German), Argument Annotated Essays 2.0 \cite{stab-gurevych-2017-parsing} and ArgRewrite V.2 \cite{kashefi2022argrewrite} corpora.
To assess the effectiveness of these revisions, we evaluate the linguistic quality of the rewritten argumentative texts both quantitatively and qualitatively.


Our contributions are as follows: (i) a comprehensive pipeline for evaluating the output quality of text rewriting tasks, consisting of 57 different metrics
(ii) a mapping of existing text generation metrics to four linguistic levels (Section \ref{sec:evaluation-setup});
(iii) a measure of what transformations are performed in a reference-based text generation setting as well as well as a measure of what grammatical changes are made, both of which are part of the text generation pipeline;
(iv) an analysis of LLM behavior on four different linguistic levels for the task of ArgImp (Section \ref{sec:results}); and (v) an investigation of LLM biases in an ArgImp setting (Section \ref{sec:length-bias}/\ref{subsec:positivity-bias}).

The code is available at \url{https://github.com/unisg-ics-dsnlp/clear-emnlp2025}.

\section{Related Work}



The capabilities of LLMs in the field of Computational Argumentation have been previously explored, particularly in the areas of Argument Mining \cite{chen-etal-2024-exploring-potential,abkenar2024assessingopensourcelargelanguage} and Argument Quality Assessment \cite{wachsmuth-etal-2024-argument,10.1007/978-3-031-63536-6_8}.
Recent work has also made use of LLMs to generate and rephrase arguments and their components. For instance, \citet{wang-etal-2025-llms-clarify} and \citet{skitalinskaya-etal-2023-claim} have used LLMs in the context of claim optimization. Moreover, \citet{ziegenbein-etal-2024-llm} present a reinforcement learning-based approach for rewriting inappropriate argumentation in online discussions. With the objective of generating complete and balanced arguments, \citet{hu-etal-2025-debate} use LLM agents to simulate a discussion among them and consolidate it into diverse and holistic arguments. 
Furthermore, \citet{hu-etal-2024-americano-argument} introduce AMERICANO, a framework with agent interaction for argument generation. It incorporates an argument refinement module that evaluates and improves argument drafts based on feedback regarding their quality.
\citet{el-baff-etal-2024-improving} make use of LLMs to rewrite existing arguments to make them more appealing to readers of a certain political ideology.



\paragraph{Argumentative Writing Support}


There is a growing line of research that focuses on developing argumentative writing support tools that provide users with feedback on the quality of their argumentative texts with the objective of guiding them in generating high-quality persuasive texts \cite{SINIKALLIO2025107598}. For instance, AL, an adaptive learning support system for argumentation skills, offers formative feedback through in-text highlighting of argumentative components, qualitative scores and graph-based visualizations of argument discourse structures \cite{wambsganss2020AL}. More recently, \citet{gubelmann2024exploring} introduced Artist, a framework that integrates LLM-based improvement suggestions. They conducted a user study with students to evaluate the effectiveness of such feedback. The results indicate that the students generally find the feedback provided by the LLMs to be helpful and of high quality. These frameworks primarily adopt a reader-oriented perspective, as their goal is to provide feedback that helps students develop the skills to refine their texts manually. Our work, in contrast, takes a text-centric approach, focusing on the linguistic quality of LLM-generated improvements in argumentative texts.




\section{Argument Improvement with LLMs}

We aim to evaluate the quality of argumentative texts improved by LLMs. Argumentation occurs in various contexts; our work centers on the following setting: (i) We focus on \textit{global argumentation} rather than local arguments. (ii) Our analysis is limited to \textit{monological texts}, excluding dialogical debates. (iii) We primarily assess \textit{intrinsic, i.e. text-focused, quality} rather than extrinsic reader-focused text effectiveness \cite{44536}.
We consider an argument $A'$ to be \textit{improved}, if $Q(A') > Q(A)$, where $Q$ is an abstract function that measures argumentative quality, $A$ is the original argument and $A'$ is an updated version of $A$ that had changes made. We explicitly do not define a universal $Q$ here, as what constitutes quality in argumentation is dependent on both intrinsic as well as extrinsic factors. An argument made by a 5-year old with grammar on the level of a high school student could generally be considered good, whereas the exact same argument, if made by a professor at a research conference, would likely be considered bad. Furthermore, an argument with wrong premises can still convince people, such as in the case of fake news, and while its intrinsic quality may be low, it can be considered successful if people believe it.

With our analysis we aim to answer the following research questions:  
(i) 
What changes on linguistic levels do LLMs make in an ArgImp setting?
(ii) 
What biases do LLMs exhibit in an ArgImp setting?
(iii) 
Do models of different sizes behave differently in an ArgImp setting?


\subsection{Model Selection}
\label{sec:model-selection}
We selected models of different families and sizes to provide a broad overview\footnote{All models are from the HuggingFace repository.}. For our experiment we used bloomz-560m and bloomz-3b \cite{muennighoff2022crosslingual}, Phi-3-mini-4k-instruct and Phi-3-medium-4k-instruct \cite{abdin2024phi}, OLMo-7B-0724-Instruct \cite{groeneveld2024olmo} and Llama-3.1-Nemotron-70B-Instruct \cite{nemotron} (henceforth referred to as Llama 3.1).

\subsection{Datasets}
Our aim is to present results on a diverse set of datasets representing different settings. We use the well-known Argument Annotated Essays 2.0 corpus by \citet{stab-gurevych-2017-parsing}, which consists of student-written essays, the Microtexts corpus \cite{peldszus2015annotated} which consists of very short argumentative texts in English and German, and the ArgRewrite V.2 corpus by \citet{kashefi2022argrewrite}, consisting of three revisions of essays by students.



Dataset metrics describing their core properties can be found in Table \ref{app:dataset-table} in Appendix \ref{app:dataset-metrics}.

\subsection{Prompting Techniques}
We use 3-shot prompting with demonstrations from the Argument Revision Corpus, Branch-Solve-Merge \citet{saha-etal-2024-branch}, Self-Discover \citet{zhou2024self}, Genetic Algorithm prompting as per \citet{guo2024connecting} and our own technique, called Little Brother, where the model is asked to give correcting feedback to an answer produced by its `little brother'. The prompts used are included in Appendix \ref{app:prompts}.

\section{Evaluation Setup}
\label{sec:evaluation-setup}

We split our evaluation into three analyses: the linguistic analysis on four linguistic levels, the bias analysis and the analysis of the argumentative discourse structure. The last one is specific to our use case of argument improvement, whereas the other analyses can be applied to all other text generation domains.

\subsection{Linguistic Analysis}

We employ a wide range of NLG evaluation metrics \cite{schmidtova-etal-2024-automatic-metrics}. Our selection aims to cover a broad spectrum of linguistic aspects to enable a comprehensive analysis of the modifications introduced by the models in our improvement setting. Following \citet{akmajian2010linguistics}, we manually mapped the metrics to their corresponding linguistic levels: 14 lexical, 22 semantic\footnote{The German sentiment scores are split into three probabilities. We consider them to be one score.}, 15 syntactic, 2 pragmatic as well as 4 argument components.



\paragraph{Lexical Analysis} 
We analyze changes on the word level as well as word distribution. We use metrics such as the number of n-syllable words and readability scores. Our aim is to provide insight into how the vocabulary the models use changes in comparison to the original texts.
We use the LexicalRichness \cite{lex} and scores from the LinguaF\footnote{\url{https://github.com/WSE-research/LinguaF}} libraries. These metrics provide an overview over the texts change on a word level. Levenshtein distance can measure how many operations it takes to transform one string into another, on a character level \cite{levenshtein1966binary}. Similar metrics are the Jaro distance \cite{Jaro01061989} and Jaro-Winkler similarity \cite{winkler1990string}.

\paragraph{Syntactic Analysis}
We expect the models to change the structure of the argumentative texts to some degree. 
To investigate structural modifications, we analyze the syntax of the sentences using dependency parse tags generated by spaCy \cite{spacy}. In that way, we aim to identify patterns in complex sentence structures and determine which types of complex clausal constructions are used more or less frequently in the improved versions of the argumentative texts.
Moreover, we make use of BERTAlign \cite{bertalign}, a sentence alignment method originally developed for the task of machine translation. It is designed to align comparable sentences from source and target languages. We use these as part of a new text generation score. We align sentences from the original texts with their corresponding improved versions. This allows us to categorize sentence transformations into several types and count their number:
(i) \textit{rephrase} and \textit{copy} (1:1); (ii) \textit{split} of an original sentence (1:$m$); (iii) \textit{merge} of original sentences ($n$:1); (iv) \textit{fusion} of original and improved sentences ($n$:$m$, where $n$ and $m$ > 1); (v) \textit{deletion} of an original sentence (1:0), and; (vi) \textit{addition} of a sentence in the improved text (0:1). This reveals what kind of modifications the models make as part of the text generation process.


\paragraph{Semantic Analysis}
To capture changes in meaning, we include a sentiment classifier, GRUEN score metrics \cite{zhu-bhat-2020-gruen}, an automated metric based on grammaticality, non-redundancy, focus, structure and coherence of texts, and a discourse analysis based on Rhetorical Structure Theory (RST) \cite{mann1988rhetorical}. We applied the parser from \citet{feng-hirst-2014-linear}. We decided against using the more recent approach by \citet{maekawa-etal-2024-obtain} due to the significantly higher computational cost and only marginal performance gains\footnote{60.0 F1 for relation classification in \citet{maekawa-etal-2024-obtain} vs. 58.2 Accuracy in \citet{feng-hirst-2014-linear}}.
We aim to capture both changes in the general tone and nuanced shifts in meaning resulting from LLMs' improvement.


\paragraph{Pragmatic Analysis}
We adopt the approach by \citet{hu-etal-2024-americano-argument} to evaluate the texts' persuasiveness and coherence as key aspects of pragmatics. The prompts we used are based on their approach and can be found in Appendix \ref{app:prompts}. The models are prompted to rate the texts with a focus on the two metrics. These metrics allow us to assess whether the improvements were successful or not, considering not only the individual changes but also the overall context of the argumentative texts. In that way, we measure the effectiveness of the communication in terms of both the texts' ability to persuade and their internal coherence within the given context.


\subsection{Bias Analysis} 
It has been discussed that LLMs have both a length\footnote{Also referred to as `verbosity bias'.} \cite{chen-etal-2024-humans,10.5555/3666122.3668142} and a positivity bias \cite{Palmer02092023,buhnila-etal-2025-chain,markowitz2024linguistic}. Length bias in this context refers to the LLM preferring longer texts. Positivity bias refers to the observation that LLM generated text tends to have a more positive tone than human-written texts. 
Verbosity bias is a relevant factor in our setting as the models may consider texts of certain lengths to be of higher quality, and perform less changes to improve them, regardless of the actual quality.
Positivity bias may cause the models to shift the tone of the argument, and could change the meaning of the argument as a whole, i.e. shifting from arguing against a topic to arguing in favor.
We investigate the presence of these biases by correlating the magnitude of changes made with the change in length as well as the sentiment of the original text.

\subsection{Analysis of the Argumentative Discourse Structure} 

We classify each sentence 
into one of the following types of argument components: claim, premise, major claim, or none, to assess structural modifications. For the English datasets, we make use of an implementation of the best-performing approach proposed in \citet{stab-gurevych-2014-identifying}
which achieves an accuracy of 0.77.
\citet{sazid-mercer-2022-unified} propose a more recent approach using deep learning models but do not significantly outperform \citet{stab-gurevych-2014-identifying}. We are not aware of any more recent approaches or available implementations as an alternative. To make the CLEAR pipeline accessible to a wider range of users we use the approach by \citet{stab-gurevych-2014-identifying}, acknowledging that higher performance may be possible by implementing a novel approach using LLMs.

For the German Microtexts, we apply the same classification, trained on the corpus introduced by \citet{wambsganss-etal-2020-corpus}, achieving an accuracy of 0.65 as reported by \citet{wambsganss2020AL}. 


\section{Results}
\label{sec:results}
Due to the large number of possible analyses\footnote{Six models, five datasets (each revision of the ArgRewrite corpus is treated as its own dataset), five prompting techniques and four linguistic levels for a total of $6 * 5 * 5 * 4 = 600$.} we focus on the most relevant combination.
The most commonly used approach for LLMs is either zero-shot or few-shot prompting, with few-shot generally performing better \cite{brown2020language}.
Based on public benchmarks, such as Chatbot Arena \cite{chiang2024chatbot}, Llama 3.1 is the best performing LLM among our selection. 
We use the combination of both Llama 3.1 as well as the 3-shot prompting approach for a deeper analysis. Both Bloomz models generated very short, barely legible texts. We omit them from the analysis. We perform a broad analysis across all other models by analyzing the scores, and further include detailed results of a manual analysis on a sample of 10 texts per dataset.

\subsection{Linguistic Analysis}

The findings are based on our proposed CLEAR pipeline. Unless otherwise stated the analysis is based on the scores of all models. Where individual models behaved differently we explicitly note this.

\begin{figure}
    \centering
    \includegraphics[width=\linewidth]{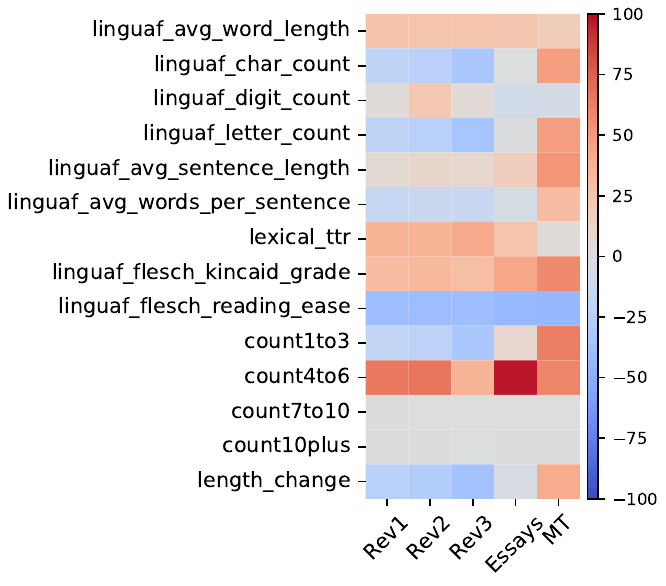}
    \caption{Changes on the lexical level for Llama 3.1. The `count' rows refer to n-syllable words. Change is measured in \%.}
    \label{fig:heatmap-lexical-pdf}
\end{figure}
\begin{figure}
    \centering
    \includegraphics[width=0.7\linewidth]{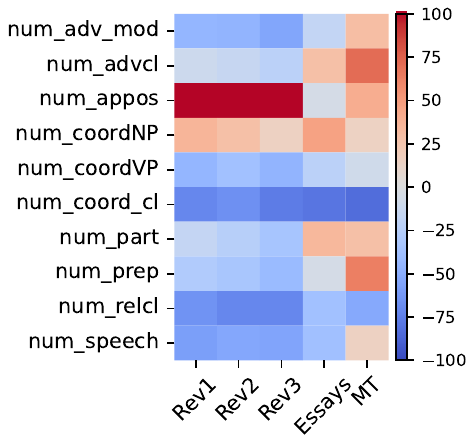}
    \caption{Changes on the syntactic level for Llama 3.1. Change is measured in \%.}
    \label{fig:heatmap-syntactic-pdf}
\end{figure}

\noindent\textbf{\textit{LLMs shorten the arguments.}} On the lexical level we note that models, on average, decreased text length ($\approx 4.66\%$ to $37.39\%$ decrease). The exception to this is the Microtext corpus, where length increased ($\approx 40.18\%$ increase). It seems that the models are aiming to add details to improve the overall quality here, as the corpus consists of very short texts.\\
\noindent\textbf{\textit{LLMs increase word length but decrease sentence length.}} Analysis on the lexical level further reveals that the models increase word lengths. We observe, particularly in the case of Llama 3.1, an increase in the number of 4 to 6 syllable words\footnote{In Figure \ref{fig:heatmap-lexical-pdf} this is labeled as `count4to6'.}, and a decrease in shorter words.\\
\noindent\textbf{\textit{LLMs reduce the reading ease.}} Larger models decreased the reading ease metrics, whereas the smaller ones increased it. This increase is not linear with the number of parameters of the models. The reason could be the increased linguistic capabilities of the larger models which make the language more complex. Manual analysis did not reveal any specific patterns that could explain this.\\
\begin{figure}
    \centering
    \includegraphics[width=0.87\linewidth]{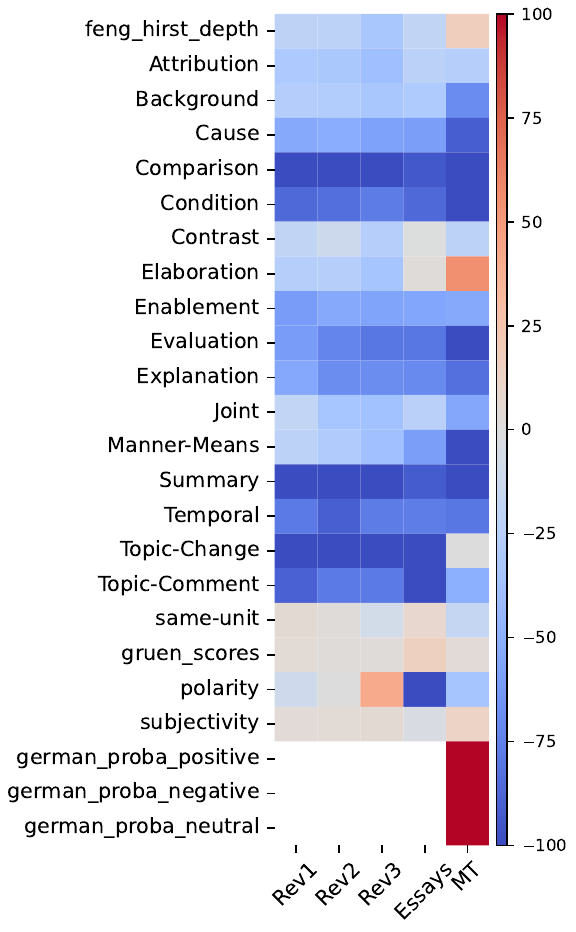}
    \caption{Changes on the semantic level for Llama 3.1. Change is measured in \%.}
    \label{fig:heatmap-semantic-pdf}
\end{figure}
{
\begin{table}[h!]
\centering
\small
\begin{tabular}{llllll}
\toprule
 & Rev1 & Rev2 & Rev3 & Essays & MT \\
\midrule
add & 0.47 & 0.44 & 0.30 & 0.36 & 0.05 \\
copy & 1.86 & 1.35 & 1.79 & 2.38 & \textbf{2.00} \\
delete & 0.31 & 0.38 & 0.74 & 0.18 & 0.03 \\
fusion & 1.02 & 0.78 & 0.81 & 0.72 & 0.26 \\
merge & \textbf{5.51} & \textbf{6.76} & \textbf{7.78} & \textbf{3.11} & 0.47 \\
other & 0.00 & 0.00 & 0.00 & 0.00 & 0.00 \\
\bottomrule
\end{tabular}
\caption{Average number of sentence transformations performed for the Llama 3.1 model. Most common transformation per dataset in bold.}
\label{table:bertalign-changes}
\end{table}
}
\noindent\textbf{\textit{LLMs transform the existing text.}} On the syntactic level the models perform operations that modify existing parts of the text (\ref{table:bertalign-changes}). 
The models rarely add entirely new sentences or paragraphs, as well as seldom entirely delete what is there.
\\
\noindent\textbf{\textit{Llama 3.1 increases the number of coordinating noun phrases.\footnote{Labelled as `num\_coordNP' in Figure \ref{fig:heatmap-syntactic-pdf}.}}} This is consistent across all the datasets. The Phi-3 models increase their number slightly, whereas OLMo consistently decreases it.\\
\noindent\textbf{\textit{Llama 3.1 significantly increases the number of appositional modifiers\footnote{Example: `The largest model, Llama 3.1, performs best.'. Here `Llama 3.1' is in apposition to `model'. Labelled as `num\_appos' in Figure \ref{fig:heatmap-syntactic-pdf}.}.}} These are commonly used to add additional details or information to other nouns or noun phrases.\\
\noindent\textbf{\textit{LLMs consistently decrease the depth of the RST parse tree.}} The analysis on the semantic level (Figure \ref{fig:heatmap-semantic-pdf}) reveals that the rhetoric structure decreases consistently across the longer corpora. A shallow RST tree indicates that the texts are less complex and easier to understand. The Microtext corpus is the exception here. 
The Microtexts are all very short arguments, and it appears as though the models consider them, or at least the overall rhetoric structure, to be too short. This is in line with prior observations that the texts themselves become shorter on all corpora but the Microtext corpus.

\noindent\textbf{\textit{LLMs make the tone more negative.}} All models perform similarly in terms of sentiment changes. On the German Microtexts the sentiment changes strongly to positive, whereas for all English texts the polarity decreases. This means the models make the texts \textit{more negative}, but not necessarily negative \textit{over all}. For Llama 3.1 we note an outlier for the polarity score on the Essays dataset. Without it, the average change in polarity is -11\%. The value for one human-written text is almost, but not quite, zero. Table \ref{tab:sentiment-original} shows the polarity scores of the original texts.\\
{
\begin{table}[h!]
\scriptsize
\begin{tabular}{@{}rrrrr|rrr@{}}
\toprule
\multicolumn{5}{r}{} & \multicolumn{3}{c}{MT (DE)} \\
Rev1 & Rev2 & Rev3 & Essays & MT (EN) & Neg & Neutral & Pos \\
\midrule
0.11 & 0.11 & 0.11 & 0.16 & 0.08 & 0.19 & 0.77 & 0.04 \\
\bottomrule
\end{tabular}
\caption{Sentiment scores of the original human-written texts in the corpora. German scores are probabilities. English ranges from -1 (negative) to +1 (positive).}
\label{tab:sentiment-original}
\end{table}
}

\begin{figure}
    \centering
    \includegraphics[width=0.9\linewidth]{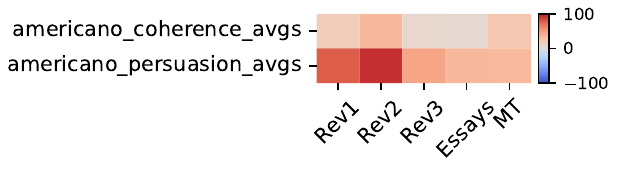}
    \caption{Changes on the pragmatic level for Llama 3.1. Change is measured in \%.}
    \label{fig:heatmap-pragmatic-pdf}
\end{figure}
\noindent\textbf{\textit{LLMs make the arguments more coherent and persuasive.}} On the pragmatic level (Figure \ref{fig:heatmap-pragmatic-pdf}) we note an increase for both persuasiveness and coherence for all models on all datasets, except for OLMo on Revision 1 of the Revisions and Microtexts datasets, where there was a small decrease for persuasion ($\approx -2.4$ and $\approx -1.8$, respectively).
Interestingly the increase in score is largest for Revision 2 for both Llama 3.1 and OLMo.
Humans wrote the original text (Revision 1) and then improved that using expert feedback to produce Revision 2. In our setting the models were asked to improve Revision 1, Revision 2 and Revision 3 separately, without this feedback. Revision 1 can be expected to be comparatively unrefined, relative to the other revisions, and as such has the most room for improvement, and Revision 3 the least. We expected the scores to decrease as revisions increase, as the texts improve with increasing revision as well.

Based on this increase in score we can say that overall the improvement process was a success. To confirm this, we have performed a manual analysis.

\subsection{Manual Analysis}

The findings in this section are based on a manual analysis of the generated texts for each dataset. We use the texts generated by Llama 3.1 for the analysis. 

\subsubsection{Analysis of Changes}
For this analysis we used 10 randomly sampled texts from each dataset. We manually compared the improved text to the original and investigated notable changes in the texts.

\noindent\textbf{\textit{Llama 3.1 mimics the style of the original text.}} We noted that if the original text had stylistic peculiarities, the model copied them. In some texts the authors include references. The model hallucinated further references that were not present in the original argument then. None of the models we used make use of Retrieval Augmented Generation (RAG), which could cause this.\\
\noindent\textbf{\textit{Llama 3.1 makes use of bullet points.}} In some cases the improved argument includes bullet points. This occurred only when the structure of the original argument lends itself well to this. Individual arguments were not well connected, such as paragraphs discussing individual claims, but not connecting to previous and following claims in neighboring paragraphs.\\
\noindent\textbf{\textit{Llama 3.1 does not appear to check for logical quality of arguments.}} In the Revisions corpus there is one text that discusses that self-driving cars can get confused by GPS trackers and drive down stairs. This is not well explained in the original texts, in any revision, but Llama 3.1 did not elaborate on it. With current technologies it is doubtful that this is a relevant factor. An improved version of the argument could either elaborate on this point or outright remove it, neither of which is something the model did.\\
\noindent\textbf{\textit{Llama 3.1 refines the existing structure.}} Often the original argument had an implicit structure in terms of paragraphs. In many cases, especially on the Essays corpus, the authors focused on one claim per paragraph. Llama 3.1 kept this structure, and often added explicit headlines to each paragraph, to illustrate what claim the paragraph is about.

\subsubsection{Analysis of Preference}
For this experiment we excluded the texts used in the analysis of the changes. We randomly sampled 20 texts from each dataset. Two reviewer with a background in computer science then blindly rated the original versus the improved argument. The order of the texts was shuffled, i.e. text 1 was not always the improved argument.

On average, the improved text was preferred 79\% of the time. Due to the small sample size of each dataset, we measured the agreement in percentage. Across all datasets, the reviewers agreed on 65.83\% of the texts.

\subsection{Length Bias}
\label{sec:length-bias}
For the correlation, we use Pearson's standard correlation coefficient. Scores are based on an analysis of Llama 3.1, OLMo and both Phi-3 models for the 3-shot prompts across all datasets. Correlations are between the original length and the delta of the metrics (original vs. improved text). We aim to analyze whether models behave differently on texts of different lengths. Strong correlations imply that the models behave differently with varying input lengths. Where p-values are omitted in the text, they are $< 0.001$.

\paragraph{Lexical}
We note a positive correlation with average word length ($\approx 0.19$) and a strong negative correlation for the 1 to 3 syllable word count ($\approx -0.49$), sentence length ($\approx -0.21$) and average words per sentence ($\approx -0.29$). The results also show a strong correlation for the token-to-type ratio ($\approx 0.46$). We find no correlation in the 7 to 10 syllable word counts and only a weak one for 10 plus counts ($\approx -0.07$). There is a positive correlation with the improved length ($\approx 0.84$). The mean length decreased from $\approx 2116$ to $\approx 1779$ characters. 

\paragraph{Syntactic}
Merge has a strong correlation ($\approx 0.64$) as well as delete ($\approx 0.27$). Both add ($\approx 0.04$, $p \approx 0.009$) and fusion ($\approx 0.05$, $p \approx 0.008$) have only a weak correlation.

\paragraph{Semantic}
We note no interesting correlations. 

\paragraph{Pragmatic}
There is no correlation ($\approx 0.013$, $p \approx 0.02$) between the length of the argumentative texts and the persuasion scores or length and coherence ($\approx -0.04$, $p \approx 0.45$).

\paragraph{Individual model behavior}
All the individual models behave similarly on the syntactic level: merge has a strong positive correlation with length ($\approx 0.55 - 0.75$ for all models), as well as delete ($\approx 0.20 - 0.32$ for all models). As these are reference-based metrics, this suggests that as text length increases, so does the number of these operations. 
In terms of the pragmatic quality dimensions of coherence and persuasion, only Llama 3.1 and Phi-3-mini show a correlation. For Llama 3.1 we observe a correlation between length and persuasion of $\approx 0.18$ while Phi-3-mini has a negative one with about $\approx 0.20$.

\paragraph{Summary}
We find a strong negative correlation for the number of 1 to 3 syllable words, as well as words per sentence and sentence length. In other words, the models tend to decrease the overall length of the texts, but do so by increasing the length of the words. The token-to-type ratio also has a strong correlation as previously discussed. This supports our hypothesis that the texts become shorter, as the words become longer: longer words are less likely to be re-used, thus increasing the types present, and a shorter text has less tokens. Both are factors leading to a higher ratio.

\subsection{Positivity Bias}
\label{subsec:positivity-bias}
We looked at the magnitude of shifts in sentiment, specifically polarity, for the Llama 3.1 model and the few-shot approach on all datasets. We measure the strength of the sentiment shifts as follows:

{
\small
\begin{equation}
    \text{shift percentage} = \left(\frac{\Delta}{|\text{Polarity Human}|}\right) * 100
\end{equation}
}
We find that $335$ negative shifts ($46.16\%$), $203$ neutral shifts ($26.40\%$) and $211$ positive shifts ($27.44\%$) occur. We consider a shift of above $+20\%$ positive, below $-20\%$ negative and between neutral. The mean is quite high with a value of $628.55\%$, but the median is negative with a value of $-14.59\%$. The mean polarity in the original texts is $+13.18\%$ and that of the improved texts is $+11.39\%$. This indicates that while positive changes are done rarely, they are strong in magnitude when they occur. Overall, the model tends to move the improved texts towards a more neutral tone.

\subsection{Argument Component Classification}
We present the changes in number of argument components in Table \ref{table:arg-components}. Components are identified on a sentence level. We note a large decrease in both non-argumentative components, as well as premises. We observe an increase in sentence length, as well as an overall merging of sentences. Due to the texts becoming shorter there can be less argument components. Despite this, we observe large decreases for the non-argumentative components, which indicates that the texts become more focused. We further hypothesize that the claims and premises are merged, as suggested by the behavior on the syntactic level, which leads to the strong decrease in the number of premises.
{
\begin{table}
\small
\begin{tabular}{llllll}
\toprule
 & Rev1 & Rev2 & Rev3 & Essays & MT \\
\midrule
MajorClaim & -0.53 & -0.50 & -0.41 & 0.26 & -0.11 \\
Claim & 0.06 & -0.16 & -0.14 & -0.42 & 1.13 \\
Premise & -4.88 & -6.55 & -8.90 & -2.54 & -0.11 \\
None & -1.65 & -2.01 & -3.26 & -0.12 & -0.89 \\
\bottomrule
\end{tabular}
\caption{Changes in number of argument components}
\label{table:arg-components}
\end{table}
}

\section{Discussion}
\label{sec:discussion}
The analysis based on the scores of our  text generation evaluation pipeline shows that on the lexical level overall text length decreases. We further observe an increase in 4 to 6 syllable words and a strong decrease in shorter words. On the syntactic level we note many merge and fuse operations, 
which means that the original text is shortened or remixed into existing sentences. Then, on the semantic level, we note a decrease in the depth of the RST parse trees. Finally, on the pragmatic level, we observe an increase in terms of coherence and persuasion, which indicates that the argument quality, in general, improved.
Manual analysis supports these empirical findings. We note that the models keep the overall structure, where it exists, and do not delete or add significant chunks of text. Instead, the models \textit{refine} and \textit{enhance} what is already there.
Notably, manual analysis revealed weak parts of certain arguments, which the model did not address or remove. 
These results together suggest that the models perform the improvement by focusing the texts:
\begin{compactitem}
    \item \textit{Lexical level}: Overall text length decreases, longer words are more common, resulting in shorter sentences composed of  longer words.
    \item \textit{Syntactic level}: Original sentences are merged, leading to shorter texts with more focused sentences.
    \item \textit{Semantic level}: Depth of the RST trees decreases, which indicates simpler texts.
    \item \textit{Pragmatic level}: Argumentative quality of the texts improves, which suggests that the models' modifications are generally effective and do not compromise the integrity of the original texts. 
    \item \textit{Manual Analysis}: Model refines the existing text, does not significantly perform changes in terms of semantics.
\end{compactitem}
In summary, it appears as though the models eliminate fluff and make the text more efficient. This is supported by our analysis of both the length and sentiment bias. To investigate the length bias we considered the token-to-type ratio as well as the lengths of the texts and sentences. The sentiment bias analyses revealed that the text shifts are towards the negative, but the original texts were positive in sentiment on average, and the improved texts are still positive, but more neutral. 


\section{Conclusion}
By making use of the CLEAR pipeline, consisting of commonly used text generation metrics mapped to linguistic levels and performing an analysis on the individuals levels,
we have found that LLMs make the texts more focused in an ArgImp setting, in the sense that (i) the texts become shorter, (ii) the average length of words increases, (iii) semantically the texts do not change. Our results suggest that the models perform well for text improvement. We note two positive factors: (i) the length of the texts decreases, but notably not in the case of the Microtexts corpus, where the input texts are already quite short, and (ii) the quality increases. 
We note small differences in model behavior in this task. Larger models performed better in both quality of the output texts and appear to make the texts more focused than the small models.
A positivity bias could not be identified, instead the models appear to aim to make the texts more neutral, instead of shifting the tone consistently to positive or negative levels. 
Lastly, we could not identify a length bias. The models do not appear to prefer texts of certain lengths. We note the tendency of Llama 3.1 in particular to use longer words, which could be a form of bias. Our results suggest that this is done to increase information density without negatively impacting readability as evident by the scores on the lexical level of our analysis.

\section{Limitations}

Our analysis focuses on textual characteristics and linguistic qualities, while disregarding more pronounced content-based aspects, overall argument quality, and reader-focused effectiveness. In particular, we do not incorporate user studies to evaluate the perceived impact of the improvements.

In the context of Automatic Essay Scoring (AES), a wide range of essay traits is typically assessed, including content, organization, word choice, sentence fluency, conventions, prompt adherence, language, narrativity, style, and voice \cite{kumar-etal-2022-many,do-etal-2023-prompt,ridley2021automated}. However, our study is limited to a narrow subset of these traits, namely text-focused linguistic qualities. Higher-order traits such as prompt adherence, content and overall organization require a more complex evaluation incorporating a detailed discourse analysis and external knowledge, which is beyond the scope of this work. By focusing on linguistic qualities, we establish a baseline for future work that may easily extend our approach to include higher-order cognitive aspects of essay quality.

Furthermore, our evaluation does not incorporate  detailed argument quality assessments grounded in argumentation theory \cite{van2013fundamentals,walton2009argumentation,mercier2011humans}. In particular, we do not account for argument quality aspects as defined by taxonomies such as the one proposed by \citet{wachsmuth-etal-2017-computational}, which extend beyond linguistic structure to include criteria such as logical soundness or dialectical reasonableness.
A recent survey by \citet{ivanova-etal-2024-lets} shows that there is no consensus regarding the different quality aspects of arguments. Varying contexts and settings make use of different metrics.
Due to the large number of existing argumentation datasets and settings in which argumentation occurs, it is not feasible to evaluate all possible metrics. This is further hindered by the fact that a majority of the metrics are not automated, lack publicly available models to score outputs automatically, do not have a sufficient amount of annotated data for model training available, or the datasets not being publicly available to begin with.

Finally, we rely largely on automatic scoring for the evaluation due to the extensive scale of our experiments. Our analysis involves five distinct datasets, six models, and five prompting techniques, each applied across four linguistic levels using 57 different metrics. This results in a total of $5 * 6 * 5 * 4 * 57 = 34'200$ combinations, thus making a manual evaluation for all combinations impractical. We included a manual evaluation on a small subset.

\bibliography{anthology,custom}

\appendix
\section{Dataset Metrics}
\label{app:dataset-metrics}
Dataset metrics can be found in Table \ref{app:dataset-table}.
{
\begin{table*}
\centering
\begin{tabular}{@{}lrrrrrr@{}}
\toprule
 & Rev1 & Rev2 & Rev3 & Essays & MT (EN) & MT (DE) \\
\midrule
Avg. Length & 3128.40 & 3464.22 & 4074.56 & 1919.51 & 452.58 & 452.58 \\
\# number of documents & 86.00 & 86.00 & 86.00 & 402.00 & 89.00 & 89.00 \\
Avg. Sentence Count & 25.42 & 28.34 & 32.76 & 16.79 & 4.18 & 4.28 \\
Avg. Sentence Length & 102.69 & 102.25 & 104.04 & 95.57 & 90.71 & 99.47 \\
Avg. Words per Sentence & 20.47 & 20.19 & 20.44 & 19.09 & 18.01 & 15.79 \\
\bottomrule
\end{tabular}
\caption{Dataset metrics of our chosen datasets.}
\label{app:dataset-table}
\end{table*}
}

\section{Prompts Used}
\label{app:prompts}
We include the prompts used here. The few-shot prompt is used for SelfDiscover as well. We otherwise follow the approach presented by \citet{zhou2024self}.
For Genetic Algorithm we use the following prompts as the initial population:

\begin{itemize}
    \item Improve the following argument
    \item Make the following argument better
    \item Enhance the following argument
    \item Make the next argument not suck
\end{itemize}

\paragraph{3-shot}
In k-shot prompting settings the model is given $k$ examples in the prompt that demonstrate the task that it should solve. Performance generally increases with larger $k$ \cite{peng-etal-2024-revisiting,10.5555/3618408.3620130}. We use demonstrations from the Argument Revision Corpus. We make use of the annotated alignment of the first and second revisions. 
Sentences for pairs of revisions are aligned and marked with the purpose. We use the first five aligned sentences that have a purpose other than `identical', for three of the essays.

\paragraph{Branch-Solve-Merge}
Branch-Solve-Merge is a prompting technique proposed by \citet{saha-etal-2024-branch}. In a first step the LLM is asked to split the problem into separate sub-problems (Branch). The sub-problems are then solved individually (Solve) and combined together into a full solution for the original problem (Merge). In our approach we ask the LLMs to come up with individual aspects that can be improved in the original argumentation (Branch). The same LLM is then prompted to improve those individual aspects (Solve) and lastly it is prompted to combine the separate generated texts into one finished argumentative text (Merge).

\paragraph{Self-Discover}
Self-Discover is a technique proposed by \citet{zhou2024self}. The LLM is first prompted to select suitable reasoning modules, from a pre-defined list, that are useful for solving the task. We use the same reasoning modules that \citet{zhou2024self} describe in their work. The model is then prompted to come up with a plan in JSON format using the modules. Finally, the plan is used to prompt the model to generate a solution. 

\paragraph{Genetic Algorithm}
A recent work by \citet{guo2024connecting} makes use of the principles of evolutionary algorithms to optimize prompts. We include an approach based on the proposed Genetic Algorithm variant. An initial prompt is used to solve the task, performance is assessed and combined with other high-performing prompts to find an optimized prompt.

\paragraph{Little Brother}
How feedback is phrased can have a large impact on how well it is received \cite{shute2008focus}. We came up with the idea to experiment with gentle feedback. The models first solve the task in the 3-shot setting, in the role of a `little brother'. Next, a `big brother' model, is asked to solve the same task, but provided the solution by the little brother model. The model is then asked to provide feedback to its `little brother'. We used Llama 3.1 as the big brother model, and the others as the solvers in the little brother role.

\begin{figure*}[!htb]
    \begin{tcolorbox}
You are given an argument about the topic "\{topic\}". Your task is to improve it. Respond only with the improved argument wrapped in @ symbols and nothing else. Here are some examples of improvements:

Demonstration1

Demonstration2

Demonstration3
    \end{tcolorbox}
    \caption{Few-shot prompt}
    \label{app:few-shot}
\end{figure*}

\begin{figure*}[!htb]
    \begin{tcolorbox}
You are given an argument about the topic >topic<. Your task is to improve it. In order to do so, your task is to first propose certain aspects of the argument that can be improved, and then divide the aspects into two groups such that the argument can be improved individually for all aspects in the groups. Your output should be in the format:

Group 1: <aspects here>

Group 2: <aspects here>
    \end{tcolorbox}
    \caption{BSM Branch prompt}
\end{figure*}

\begin{figure*}[!htb]
    \begin{tcolorbox}
Improve the following argument by focussing on the specific aspects. Respond with the improved argument wrapped in @ symbols. Try to keep the length of the improved argument similar to the original one.

Argument: >task<

Aspects: >group<
    \end{tcolorbox}
    \caption{BSM Solve prompt}
\end{figure*}

\begin{figure*}[!htb]
    \begin{tcolorbox}
Given two arguments about the topic >topic<, your task is to merge them into a single argument. Respond with the merged argument wrapped in @ symbols.
    \end{tcolorbox}
    \caption{BSM Merge prompt}
\end{figure*}

\begin{figure*}[!htb]
    \begin{tcolorbox}
You are given two arguments. Your task is to choose the better one. Respond with @First@ if you prefer the first one, and with @Second@ if you prefer the second one.
    \end{tcolorbox}
    \caption{Genetic Algorithm population scoring prompt}
\end{figure*}

\begin{figure*}[!htb]
    \begin{tcolorbox}
Solve this task: {task}. Your little brother has solved this task like this previously:

[PREVIOUS]

\{previous\}

[/PREVIOUS]

Check if your little brother's solution is correct. If it is not, teach them where they made a mistake, and correct it. If it is correct, state the solution and explain it. Put the corrected solution into @ symbols.
    \end{tcolorbox}
    \caption{Little Brother prompt}
\end{figure*}

\begin{figure*}[!htb]
    \begin{tcolorbox}
You are a lecturer of the writing class. You are given the following proposition on a controversial topic. You need to carefully read the proposition and evaluate it based on the criteria:

- Clarity

- Relevance

- Logical consistency

- Validity of reasoning

Now you need to assign a score for coherence on a scale of 1 to 5, where 1 is the lowest and 5 is the highest based on the Evaluation Criteria. Note, you should be very strict when giving the score.
    \end{tcolorbox}
    \caption{AMERICANO coherence prompt}
\end{figure*}

\begin{figure*}[!htb]
    \begin{tcolorbox}
You are a lecturer of the writing class. You are given the following proposition on a controversial topic. You need to carefully read the proposition and evaluate it based on the criteria:

- Language and rhetoric

- Addressing opposing viewpoints

- Credibility

- Overall effectiveness

Now you need to assign a score for persuasion on a scale of 1 to 5, where 1 is the lowest and 5 is the highest based on the Evaluation Criteria. Note, you should be very strict when giving the score.
    \end{tcolorbox}
    \caption{AMERICANO persuasion prompt}
\end{figure*}

\clearpage
\onecolumn
\subsection{Reasoning Modules}
We include the reasoning modules used for the Self-Discover prompting here. They are same ones as used in \cite{zhou2024self}.

\begin{itemize}
\item How could I devise an experiment to help solve that problem?
\item Make a list of ideas for solving this problem, and apply them one by one to the problem to see if any progress can be made.
\item How could I measure progress on this problem?
\item How can I simplify the problem so that it is easier to solve?
\item What are the key assumptions underlying this problem?
\item What are the potential risks and drawbacks of each solution?
\item What are the alternative perspectives or viewpoints on this problem?
\item What are the long-term implications of this problem and its solutions?
\item How can I break down this problem into smaller, more manageable parts?
\item Critical Thinking: This style involves analyzing the problem from different perspectives, questioning assumptions, and evaluating the evidence or information available. It focuses on logical reasoning, evidence-based decision-making, and identifying potential biases or flaws in thinking.
\item Try creative thinking, generate innovative and out-of-the-box ideas to solve the problem. Explore unconventional solutions, thinking beyond traditional boundaries, and encouraging imagination and originality.
\item Seek input and collaboration from others to solve the problem. Emphasize teamwork, open communication, and leveraging the diverse perspectives and expertise of a group to come up with effective solutions.
\item Use systems thinking: Consider the problem as part of a larger system and understanding the interconnectedness of various elements. Focuses on identifying the underlying causes, feedback loops, and interdependencies that influence the problem, and developing holistic solutions that address the system as a whole.
\item Use Risk Analysis: Evaluate potential risks, uncertainties, and tradeoffs associated with different solutions or approaches to a problem. Emphasize assessing the potential consequences and likelihood of success or failure, and making informed decisions based on a balanced analysis of risks and benefits.
\item Use Reflective Thinking: Step back from the problem, take the time for introspection and self-reflection. Examine personal biases, assumptions, and mental models that may influence problem-solving, and being open to learning from past experiences to improve future approaches.
\item What is the core issue or problem that needs to be addressed?
\item What are the underlying causes or factors contributing to the problem?
\item Are there any potential solutions or strategies that have been tried before? If yes, what were the outcomes and lessons learned?
\item What are the potential obstacles or challenges that might arise in solving this problem?
\item Are there any relevant data or information that can provide insights into the problem? If yes, what data sources are available, and how can they be analyzed?
\item Are there any stakeholders or individuals who are directly affected by the problem? What are their perspectives and needs?
\item What resources (financial, human, technological, etc.) are needed to tackle the problem effectively?
\item How can progress or success in solving the problem be measured or evaluated?
\item What indicators or metrics can be used?
\item Is the problem a technical or practical one that requires a specific expertise or skill set? Or is it more of a conceptual or theoretical problem?
\item Does the problem involve a physical constraint, such as limited resources, infrastructure, or space?
\item Is the problem related to human behavior, such as a social, cultural, or psychological issue?
\item Does the problem involve decision-making or planning, where choices need to be made under uncertainty or with competing objectives?
\item Is the problem an analytical one that requires data analysis, modeling, or optimization techniques?
\item Is the problem a design challenge that requires creative solutions and innovation?
\item Does the problem require addressing systemic or structural issues rather than just individual instances?
\item Is the problem time-sensitive or urgent, requiring immediate attention and action?
\item What kinds of solution typically are produced for this kind of problem specification?
\item Given the problem specification and the current best solution, have a guess about other possible solutions.
\item Let’s imagine the current best solution is totally wrong, what other ways are there to think about the problem specification?
\item What is the best way to modify this current best solution, given what you know about these kinds of problem specification?
\item Ignoring the current best solution, create an entirely new solution to the problem.
\item Let’s think step by step.
\item Let’s make a step by step plan and implement it with good notion and explanation.
\end{itemize}
\twocolumn
\clearpage

\section{Scores}
\label{app:scores}
The following tables show the scores of the Llama 3.1 model with the 3-shot prompting approach. We omit the other tables due to the large amount of data. Scores for all models and approaches are included in the Github repository.

\begin{table*}[t]
\small
\centering
\begin{tabular}{llllll}
\toprule
index & Rev1 & Rev2 & Rev3 & Essays & MT \\
\midrule
add & 46.51 & 44.19 & 30.23 & 36.32 & 5.06 \\
copy & 186.05 & 134.88 & 179.07 & 237.56 & 200.00 \\
delete & 31.40 & 38.37 & 74.42 & 17.91 & 2.81 \\
fusion & 102.33 & 77.91 & 81.40 & 71.89 & 26.40 \\
merge & 551.16 & 675.58 & 777.91 & 311.44 & 47.19 \\
other & 0.00 & 0.00 & 0.00 & 0.00 & 0.00 \\
\bottomrule
\end{tabular}
\caption{BERTAlign changes}
\end{table*}

\begin{table*}
\small
\centering
\begin{tabular}{llllll}
\toprule
score\_name & Rev1 & Rev2 & Rev3 & Essays & MT \\
\midrule
linguaf\_avg\_word\_length & 26.26 & 25.43 & 25.01 & 23.58 & 17.77 \\
linguaf\_char\_count & -20.51 & -23.48 & -33.29 & -0.28 & 47.54 \\
linguaf\_digit\_count & 2.53 & 22.66 & 3.43 & -8.09 & -7.05 \\
linguaf\_letter\_count & -21.74 & -24.76 & -34.45 & -1.06 & 47.46 \\
linguaf\_avg\_sentence\_length & 5.30 & 8.67 & 7.30 & 16.99 & 52.27 \\
linguaf\_avg\_words\_per\_sentence & -16.59 & -13.43 & -14.11 & -5.24 & 30.39 \\
lexical\_ttr & 35.32 & 35.55 & 40.82 & 25.40 & 1.73 \\
linguaf\_flesch\_kincaid\_grade & 31.07 & 32.12 & 29.37 & 42.29 & 57.07 \\
linguaf\_flesch\_reading\_ease & -40.96 & -41.29 & -40.62 & -43.37 & -44.43 \\
original\_length & 312839.53 & 346422.09 & 407455.81 & 191951.49 & 47249.44 \\
count1to3 & -18.61 & -22.22 & -32.49 & 7.86 & 61.73 \\
count4to6 & 64.32 & 65.80 & 36.06 & 96.24 & 58.65 \\
count7to10 & 0.00 & -0.41 & -0.40 & -0.21 & -0.39 \\
count10plus & 0.00 & 0.00 & -0.39 & 0.00 & 0.00 \\
length\_change & -24.95 & -27.48 & -37.39 & -4.66 & 40.18 \\
levenshtein\_levenshtein & 2045.71 & 2265.07 & 2687.08 & 1287.39 & 407.94 \\
\bottomrule
\end{tabular}
\caption{Lexical Level}
\end{table*}

\begin{table*}
\small
\centering
\begin{tabular}{llllll}
\toprule
score\_name & Rev1 & Rev2 & Rev3 & Essays & MT \\
\midrule
add & 46.51 & 44.19 & 30.23 & 36.32 & 5.06 \\
copy & 186.05 & 134.88 & 179.07 & 237.56 & 200.00 \\
delete & 31.40 & 38.37 & 74.42 & 17.91 & 2.81 \\
fusion & 102.33 & 77.91 & 81.40 & 71.89 & 26.40 \\
merge & 551.16 & 675.58 & 777.91 & 311.44 & 47.19 \\
other & 0.00 & 0.00 & 0.00 & 0.00 & 0.00 \\
num\_adv\_mod & -45.75 & -47.55 & -56.07 & -18.38 & 29.56 \\
num\_advcl & -12.24 & -15.63 & -23.12 & 28.10 & 70.23 \\
num\_appos & 132.15 & 137.54 & 151.76 & -6.37 & 39.52 \\
num\_coordNP & 35.08 & 28.10 & 14.49 & 46.06 & 13.69 \\
num\_coordVP & -45.84 & -38.28 & -47.45 & -23.03 & -9.93 \\
num\_coord\_cl & -72.38 & -67.23 & -77.94 & -81.12 & -84.47 \\
num\_part & -17.53 & -25.56 & -35.03 & 33.06 & 27.43 \\
num\_prep & -29.26 & -33.67 & -42.91 & -6.82 & 62.17 \\
num\_relcl & -65.74 & -72.16 & -72.42 & -38.11 & -53.40 \\
num\_speech & -59.85 & -55.08 & -56.76 & -39.01 & 14.29 \\
improved\_length & 2347.79 & 2512.23 & 2550.88 & 1830.12 & 662.37 \\
original\_length & 3128.40 & 3464.22 & 4074.56 & 1919.51 & 472.49 \\
\bottomrule
\end{tabular}
\caption{Syntactic Level}
\end{table*}

\begin{table*}
\small
\centering
\begin{tabular}{llllll}
\toprule
score\_name & Rev1 & Rev2 & Rev3 & Essays & MT \\
\midrule
feng\_hirst\_depth & -21.17 & -22.09 & -33.26 & -19.39 & 17.15 \\
Attribution & -31.49 & -32.23 & -39.57 & -23.15 & -26.27 \\
Background & -27.05 & -28.40 & -33.39 & -29.81 & -70.00 \\
Cause & -53.88 & -51.09 & -57.85 & -59.51 & -91.67 \\
Comparison & -100.00 & -100.00 & -100.00 & -94.12 & -100.00 \\
Condition & -86.60 & -84.09 & -77.82 & -86.11 & -100.00 \\
Contrast & -19.13 & -11.90 & -25.84 & -0.24 & -22.45 \\
Elaboration & -26.43 & -26.56 & -35.18 & 2.19 & 54.79 \\
Enablement & -61.29 & -53.89 & -56.55 & -55.63 & -53.85 \\
Evaluation & -60.98 & -72.97 & -79.81 & -79.78 & -100.00 \\
Explanation & -54.65 & -68.97 & -69.29 & -70.40 & -83.33 \\
Joint & -18.71 & -34.98 & -38.28 & -24.17 & -55.02 \\
Manner-Means & -22.55 & -29.05 & -38.89 & -59.72 & -100.00 \\
Summary & -100.00 & -100.00 & -100.00 & -92.31 & -100.00 \\
Temporal & -78.33 & -90.74 & -77.35 & -76.77 & -80.00 \\
Topic-Change & -100.00 & -100.00 & -100.00 & -100.00 & 0.00 \\
Topic-Comment & -90.22 & -78.57 & -78.57 & -100.00 & -50.00 \\
same-unit & 5.10 & 2.01 & -8.18 & 7.97 & -16.89 \\
gruen\_scores & 4.02 & 2.28 & 1.94 & 15.11 & 3.38 \\
polarity & -10.53 & 0.54 & 40.92 & -1157.79 & -35.83 \\
subjectivity & 3.60 & 4.50 & 4.90 & -3.29 & 13.20 \\
german\_proba\_positive & nan & nan & nan & nan & 162.27 \\
german\_proba\_negative & nan & nan & nan & nan & 107.08 \\
german\_proba\_neutral & nan & nan & nan & nan & 146.89 \\
\bottomrule
\end{tabular}
\caption{Semantic Table}
\end{table*}

\begin{table*}
\small
\centering
\begin{tabular}{llllll}
\toprule
score\_name & Rev1 & Rev2 & Rev3 & Essays & MT \\
\midrule
americano\_coherence\_avgs & 18.13 & 32.60 & 8.35 & 6.45 & 23.11 \\
americano\_persuasion\_avgs & 76.18 & 91.32 & 44.00 & 32.52 & 31.31 \\
\bottomrule
\end{tabular}
\caption{Pragmatic Level}
\end{table*}

\begin{table*}
\small
\centering
\begin{tabular}{llllll}
\toprule
dataset & Rev1 & Rev2 & Rev3 & Essays & MT \\
\midrule
Claim & 0.06 & -0.16 & -0.14 & -0.42 & 1.13 \\
MajorClaim & -0.53 & -0.50 & -0.41 & 0.26 & -0.11 \\
None & -1.65 & -2.01 & -3.26 & -0.12 & -0.89 \\
Premise & -4.88 & -6.55 & -8.90 & -2.54 & -0.11 \\
\bottomrule
\end{tabular}
\caption{Argument Mining Components}
\end{table*}

\section{License Terms of Used Datasets}
We used the Argument Annotated Essays 2.0 \cite{stab-gurevych-2017-parsing} in our research. This dataset may only be used for academic and research purposes.

The ArgRewrite V.2 \cite{kashefi2022argrewrite} corpus is available under the GNU General Public license.

The Microtexts corpus \cite{peldszus2015annotated} is available under a Creative Commons Attribution-NonCommercial-ShareAlike 4.0 International License.

\section{Computational details}
We used the following models for our experiments:
\begin{itemize}
    \item bigscience/bloomz-3b
    \item bigscience/bloomz-560
    \item allenai/OLMo-7B-0724-Instruct-hf
    \item microsoft/Phi-3-medium-4k-instruct (14B parameters)
    \item microsoft/Phi-3-mini-4k-instruct (3.8B parameters)
    \item nvidia/Llama-3.1-Nemotron-70B-Instruct
\end{itemize}

All models are from the HuggingFace repository.

Our texts were generated on up to 8 V100 GPUs on a DGX2 machine over the course of four weeks. Experiments were performed consecutively and did not run the full four weeks. Llama 3.1 is the only model that needed eight GPUs, the other models ran on up to four GPUs if resources were available, but can be run on two. Total GPU hours for both text generation and scoring are around $\approx 20$.

\section{Use of AI assistants}
We used ChatGPT 4o to generate the title of the paper.

\section{Annotation Details}
Both annotators are authors of the paper and were aware that their annotations would be used as part of this paper. They are Caucasian and from Central Europe. The instructions were to analyze the changes that are present in the improved texts and to choose the argument that they consider to be better in the preference analysis.

\end{document}